\documentclass[a4paper, 10pt, conference]{ieeeconf}
\IEEEoverridecommandlockouts
\usepackage{multirow}
\usepackage{cite}
\usepackage{amsmath,amssymb,amsfonts}
\usepackage[noend]{algorithmic}
\usepackage{graphicx}
\usepackage{textcomp}
\usepackage{xcolor}
\usepackage{algorithm}
\usepackage{hyperref}
\usepackage{bm}

\def\BibTeX{{\rm B\kern-.05em{\sc i\kern-.025em b}\kern-.08em
    T\kern-.1667em\lower.7ex\hbox{E}\kern-.125emX}}
    
\hypersetup{
    colorlinks=true,
    linkcolor=blue,
    filecolor=magenta,      
    urlcolor=blue,
}

\title{\LARGE \bf FAITH: Fast iterative half-plane focus of expansion estimation\\ using event-based optic flow}

\author{Raoul Dinaux, Nikhil Wessendorp, Julien Dupeyroux and Guido C.~H.~E. de Croon\authorrefmark{1}

\thanks{\authorrefmark{1}All authors are with Faculty of Aerospace Engineering,  Delft  University  of  Technology,
        Kluyverweg 1, 2629HS Delft, The Netherlands.
        {\tt\small j.j.g.dupeyroux@tudelft.nl}}
}

\begin{document}
\maketitle

\begin{abstract}
Course estimation is a key component for the development of autonomous navigation systems for robots. While state-of-the-art methods widely use visual-based algorithms, it is worth noting that they all fail to deal with the complexity of the real world by being computationally greedy and sometimes too slow. They often require obstacles to be highly textured to improve the overall performance, particularly when the obstacle is located within the focus~of~expansion (FOE) where the optic~flow (OF) is almost null. This study proposes the \hbox{FAst~ITerative~Half-plane (FAITH)} method to determine the course of a micro~air~vehicle (MAV). This is achieved by means of an event-based camera, along with a fast RANSAC-based algorithm that uses event-based OF to determine the FOE. The performance is validated by means of a benchmark on a simulated environment and then tested on a dataset collected for indoor obstacle avoidance. Our results show that the computational efficiency of our solution outperforms state-of-the-art methods while keeping a high level of accuracy. This has been further demonstrated onboard an MAV equipped with an event-based camera, showing that our event-based FOE estimation can be achieved online onboard tiny drones, thus opening the path towards fully neuromorphic solutions for autonomous obstacle avoidance and navigation onboard MAVs.
\end{abstract}

\section{Introduction}
\label{sec:introduction}

Autonomous navigation, including path planning, obstacle avoidance, and localization, for both ground and aerial robots is considered as one of the top ten technological challenges of our time~\cite{yang2018grand}. Despite outstanding studies in this field, it must be noticed that we still fail at tackling real-world scenarios, where lighting conditions can change abruptly, light can be absent, and where obstacles can severely hamper the performance of the navigation system running onboard the robot. Another crucial aspect for making an autonomous navigation system suitable for real-world applications is to make sure that the robot can deal with high speeds. This is precisely one of the bottlenecks for drone applications, where computational resources and energy usage are important factors for the viability of the proposed method. Lastly, the navigation system must be endowed with deep auto-adaptation skills to make it worth deploying onboard robots in complex environments that remain hard to model, or of which the core nature is simply not understood yet. 

The limitations of navigation systems are multi-factorial, but the most important reason for this may be the sensing component itself. While observing the animal kingdom, one can note that each species optimized its sensors to better evolve in its environment. For instance, birds and insects are sensitive to the polarization state of the skylight to estimate their course, dung beetles retrieve navigational cues from the Milky Way, and eagles have an extremely high visual acuity to better find their preys. In comparison, robots are often equipped with cameras for which both the temporal ($30-60~fps$ on average) and the visual resolutions are limited. This gets even more crucial with small drones. Given their small dimensions and weight, micro air vehicles (MAVs) are safe to operate autonomously around humans in complex environments. Unfortunately, MAVs are endowed with highly restricted power capacity, and extremely limited computational resources. Their use is also hampered by the risk of GPS failure indoor, magnetometer disturbances because of surrounding ferrous materials (buildings, infrastructures), and IMU drift over time. Embedded cameras have greatly contributed to the reduction of navigation failure, but their use remains limited by the low computational resources available onboard. It is therefore crucial to determine fast and efficient methods to allow MAVs to autonomously navigate and avoid both static and moving obstacles.

\begin{figure}
    \centering
    \includegraphics[width=1\linewidth]{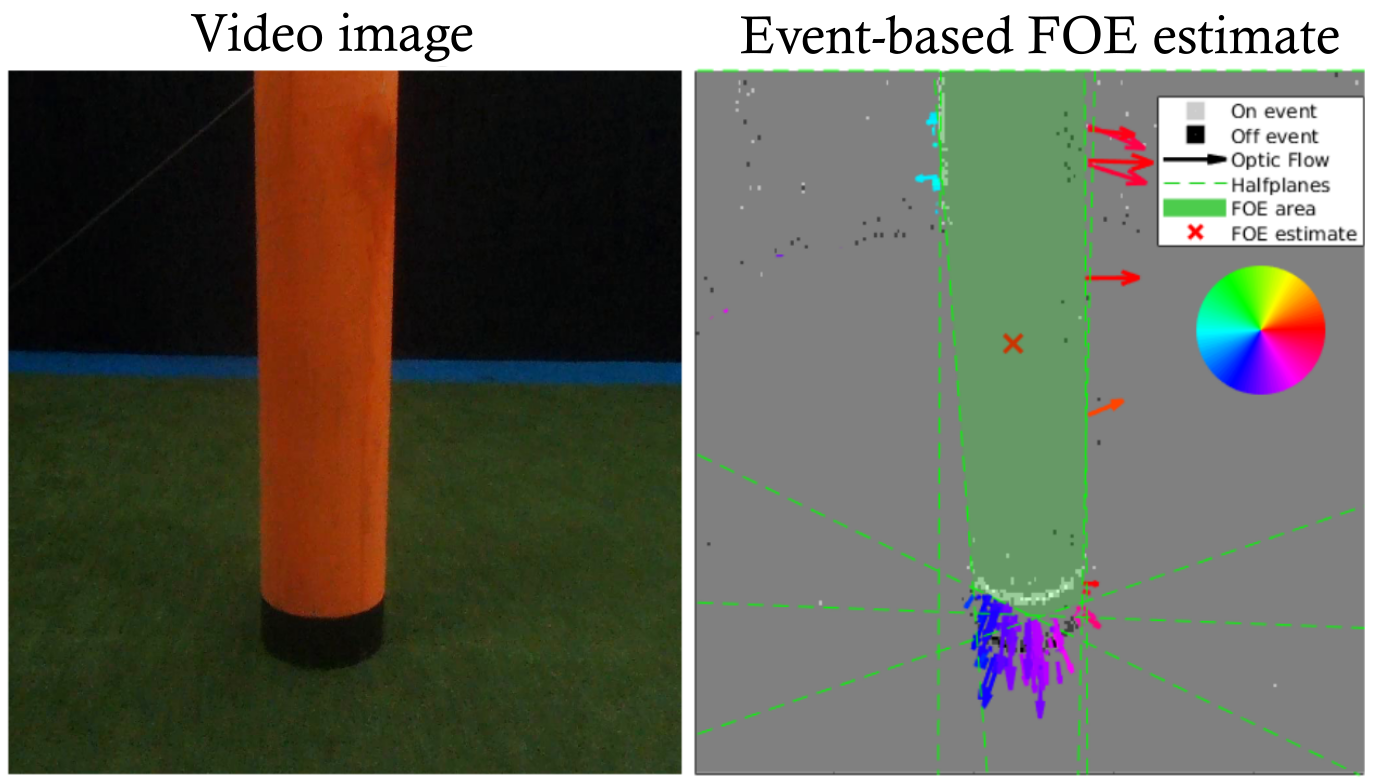}
    \caption{An application of the FAITH method for fast and accurate FOE estimation in MAV flight towards a single pole. The left plot shows a video-image from the MAV, flying towards a single pole in the TU Delft flying arena. The right plot shows an event-image of the pole with an overlay of optic flow vectors and the FOE estimation, performed by our method.}
    \label{fig:FOEexample}
\end{figure}

The recent developments in neuromorphic systems represent a promising opportunity for autonomous obstacle avoidance and navigation onboard robots, in particular for MAVs. In this respect, event-based cameras were first released in 2008 by Lichtensteiner et al.~\cite{patrick2008128x}. Unlike conventional cameras which output images at a fixed frame-rate, event-based cameras produce a stream of asynchronous and independent events reporting changes in brightness at the pixel level~\cite{Gallego2020event}. Therefore, these cameras inherently capture the apparent motion. The intensity change threshold, which triggers the pixel, is user-defined. The events are labeled by the pixel location, trigger time and a polarity ($+1$ for positive change of brightness, $-1$ for a negative change). Event-based cameras offer a high dynamic range ($>120~dB$) along with a high temporal resolution (in the range of microseconds). These advantages make event-based cameras inherently insensitive to classical visual artifacts such as motion blur or the tunnel effect. As a result, these cameras provide accurate visual information at extremely high speed, making them suitable for aerial robotics, including MAVs, for various tasks such as obstacle avoidance~\cite{falanga2020dynamic, mitrokhin2018event} and visual odometry~\cite{vidal2018ultimate}. 

In this study, we propose the FAITH (FAst ITerative Half-plane) method to estimate the course of the MAV by means of an event-based camera (i.e., the DVS240C~\cite{brandli2014240}), along with a fast RANSAC-based algorithm for the determination of the focus of expansion (FOE) using optic flow (OF) as an input (Fig.\ref{fig:FOEexample}). Optic flow is described as the pattern of apparent motion of objects in a visual scene caused by the relative motion between the observer and a scene~\cite{gibson1955parallax}. The FOE is therefore defined as the singular point from which the apparent OF expands, assuming the scene is static and the motion of the observer is purely translational. This point indicates the course of the observer, and therefore is a crucial element in visual-based navigation. Appendix~\ref{app:OFtheory} gives a theoretical background to OF and FOE estimation. Determining the FOE is challenging as only normal flow is available, and the computational limitation of the MAV does not allow for expensive online visual-processing. 

The determination of the FOE onboard mobile systems equipped with cameras has received large attention from researchers over the past decades, showing a great variety of approaches to solve this very complex problem. In the following study, we focused on sparse OF-based FOE estimation, for which state-of-the-art solutions currently available can be divided into three categories: (i) counting vectors directions~\cite{Souhila2007Optical,Huang2018AnEW}, (ii) creating a probability map based on negative vector intersections and (iii) based on negative half-planes. 

Methods relying on counting vectors show limited performance when exploited in online MAVs application. To reduce the computation cost of online FOE estimation, methods based on probability maps seem to be a promising alternative. Guzel et al.~\cite{Guzel2010Optical} proposed to compute a probability map based on the amount of OF vector intersections per location. They demonstrated the performance of their method through a navigation task with a ground robot equipped with a camera. A similar method was implemented by Buczko et al.~\cite{Buczko2017Monocular}, where RANSAC scheme randomly selects two OF vectors, create a candidate FOE location by calculating the intersection, and test this location against all OF vectors. After a predetermined amount of iterations, the candidate with the highest amount of inliers is selected as the FOE. Results obtained with a RGB camera showed a translation error as low as 0.81\%. Yet, using vectors intersection remains a limited solution to FOE estimation since OF estimation on natural scenes is a complex task and the resulting estimates (normal flow) can differ from the true flow.

To compensate for this, it has been proposed to build the probability map using the negative half-planes~\cite{Clady2014AsynchronousVE}. As the normal flow is computed, the assumption is made that the FOE must lie in the negative half-plane of as many normal OF vectors as possible. For each OF vector an orthogonal line is taken, which intersects the vector location. The negative half-plane of this orthogonal line is used to update the probability map. All locations which are not updated are subject to exponential decay over time. The location with the highest value on the probability map is selected as the FOE. This method has been used with event-based cameras to estimate time-to-contact (TTC) in the context of obstacle avoidance with MAVs~\cite{Colonnier2018ObstacleAU}. 

Although the negative half-planes approach suggest an improvement in the course estimation, it is worth noting that the new computation introduced in the plane estimation and intersection considerably affect the overall performance. 

\noindent\textbf{Contributions }-- We propose the FAITH method for FOE estimation based on negative half-planes intersections, further optimized by means of a RANSAC process. Our contributions are:\begin{enumerate}
    \item[(a)] a novel course estimation algorithm (FAITH) that is highly computationally efficient, runs real-time onboard robots (including MAVs), and provides a robust estimate of the FOE even with poor-textured obstacles;
    \item[(b)] an exhaustive assessment of the overall performance of the FAITH algorithm, first using the ESIM event-based camera simulator~\cite{rebecq2018esim}, and then using an extensive dataset collected in the TU Delft flying arena equipped with the OptiTrack motion tracking system;
    \item[(c)] a real-world demonstration of the performance onboard an MAV designed for the purpose of this study.
\end{enumerate} 

\section{Materials and methods}
\subsection{The proposed FAITH method}

\begin{figure}[!t]
    \centering
    \includegraphics[width=0.35\textwidth]{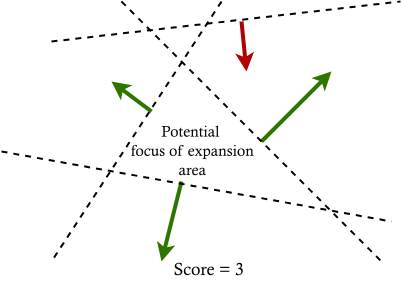}
    \setlength{\belowcaptionskip}{-5pt}
    \caption{Schematic example of an FOE estimation by the FAITH method. The arrows represent normal optic flow, the dotted lines their orthogonal half-planes. As the centre of the potential FOE area lies within three half-planes, the iteration score of this estimation is three.}
    \label{fig:halfplane}
\end{figure}

We apply an event-surface method to compute the local normal flow based on visual events streamed by an even-based camera~\cite{Benosman2014event,Hordijk2018vertical}. When an OF vector is available, we assume the FOE lies in the negative half-plane delimited by the straight line orthogonal to the OF vector (Fig.~\ref{fig:halfplane}). The aperture problem limits OF on edges to be normal to the edge, whereas only OF on corners result in true OF. Therefore, the assumption is made that the FOE must lie in the negative half-plane of a line orthogonal to the OF vector. We then build on the approach from~\cite{Clady2014AsynchronousVE} to compute a probability region for the FOE estimation. As MAVs are limited by their computational resources, the ego-motion estimation should require as little as computation as possible while still assuring accuracy. Because we are using an event-based camera, the algorithm implemented in~\cite{Clady2014AsynchronousVE}, which updates all pixels of the probability map for each OF vector, will inevitably lead to large computational needs. 

To compensate for the computational cost, we propose to apply a RANSAC scheme to create an FOE area by taking the intersection of the negative half-planes of two randomly chosen vectors (Algorithm~\ref{alg:foe}). A new OF vector is then chosen and the intersection of the negative half-plane is updated. If the new vector reduces the size of the FOE area, it is added as a new boundary. This process is continued until the new chosen vector does not reduce the size of the FOE area. Then the center of this area is calculated (Fig.~\ref{fig:halfplane}) and an iteration score is assigned by computing in how many negative half-planes the FOE estimate lies. The score and center position are saved and another iteration is performed. After a user-defined amount of iterations, the search is stopped and the iteration with the highest score is chosen as the best FOE candidate.

\begin{algorithm}[!t]
    \caption{FAITH method for FOE estimation}
    \begin{algorithmic}
        \FOR{\textit{iterations}}
        \STATE $ \textit{stop search} = \text{false}$
        \STATE Pick two random OF vectors.
        \STATE Calculate $\textit{current FOE area}$ (bounded by negative half-planes orthogonal to the selected vectors).
        \WHILE{$ \textit{stop search} == \text{false}$}
        \STATE Pick new random vector.
        \STATE Calculate \textit{new area} (bounded by $\textit{current FOE area}$ and the negative half-plane of the selected vector).
        \IF {$ \textit{new area} < \textit{current FOE area}$}
        \STATE $\textit{current FOE area} = \textit{new area}$ 
        \ELSE
        \STATE Calculate \textit{score} as the total amount half-planes the center of \textit{new area} lies in.
        \STATE \textit{stop search} $=$ true
        \ENDIF
        \IF{$ \textit{score} > \textit{max score}$}
        \STATE \textit{max score} $=$ \textit{score}
        \STATE \textit{best area} $=$ \textit{new area}
        \ENDIF
        \ENDWHILE
        \ENDFOR
        \STATE FOE $=$ center of \textit{best area}
    \end{algorithmic}
    \label{alg:foe}
\end{algorithm}

\subsection{Computational complexity analysis}
\label{sec:compcomplexity}

Given that our proposed method extends the one introduced in Clady et al.~\cite{Clady2014AsynchronousVE}, we determined the computational complexity of both algorithms to assess their overall computational performances. For each vector, the method by Clady et al. updates all locations of a probability map. Therefore the computational complexity for this method can be written as $\mathcal{O}(N * M_p)$, with $N$ the number of OF vectors and $M_p$ the number of pixels in the probability map. 

In our method, the majority of the computational complexity lies in checking how many inliers the candidate locations have. This depends on the total of OF vectors and candidate locations, which is equal to the user-defined number of iterations run. Therefore it can be expressed as $\mathcal{O}(N * I)$, with $N$ the number of OF vectors and $I$ the number of iterations (i.e., potential FOE locations). To get an estimate of this $I$, the theoretical minimum of RANSAC iterations required to construct a proper model with a chosen probability is:

\begin{equation}
I = \frac{log(1-p)}{log(1-w^n)}
\label{eq:RANSAC_its}
\end{equation}

\noindent where $I$ is the required number of iterations, $p$ is the probability of selecting a proper model, $w$ is the ratio between inliers and the total set, and $n$ is the sum of inliers required for a proper model.

This formula can be seen as a theoretical upper bound as it assumes that the random selection of vectors can include already chosen vectors. For example: requiring a probability of 95\% to find a proper model, assuming at least 10 vectors are required for creating a proper model and assuming that 75\% of the total set consists of inliers, the total amount of iterations required is 52. Comparing the computational complexity of both methods (assuming a $240\times180$ pixel probability map) shows that the method implemented by Clady et al., $\mathcal{O}(43200 * N)$, is a few orders of magnitude more complex than the proposed method, $\mathcal{O}(52 * N)$.  Therefore, it is concluded that in the general case the theoretical computational complexity of our method is lower, and user-defined.

\section{Performance benchmark}
\label{sec:benchmark}
To assess the performance of our method, we first test it in a virtual environment featuring an event-based camera simulator. Then, we demonstrate its robustness by testing it on a manually controlled obstacle avoidance dataset that we collected in our indoor flying arena equipped with the OptiTrack motion capture system (see Supplementary Materials). Lastly, the online performance is demonstrated by testing the method onboard an MAV equipped with a DVS240 camera in an autonomous obstacle detection and avoidance task. For the sake of experimental simplification, the motion of the MAV is bounded to translation in the horizontal plane and within the camera FOV. Appendix~\ref{app:FOEoutside} discusses the impact this assumption. While assessing the performance of our proposed method, we compare it with the state-of-the-art FOE estimation methods. As described in Section~\ref{sec:introduction}, three categories of FOE estimation methods using sparse normal flow are identified. Therefore, we also implement the three following algorithms to test them on both simulated and real-world dataset: (i) the vector counting method from Huang et al.~\cite{Huang2018AnEW}, (ii) the probability map method based on vector intersections implemented by Buczko~et~al.~\cite{Buczko2017Monocular}, and (iii) the negative half-planes method introduced by Clady~et~al.~\cite{Clady2014AsynchronousVE}, further referred to as 'NESW', 'Vec. Intersections', and 'Half-planes' respectively.

\subsection{Benchmark on simulated data}

\begin{figure}[!t]
    \centering
    \includegraphics[width=0.9\linewidth]{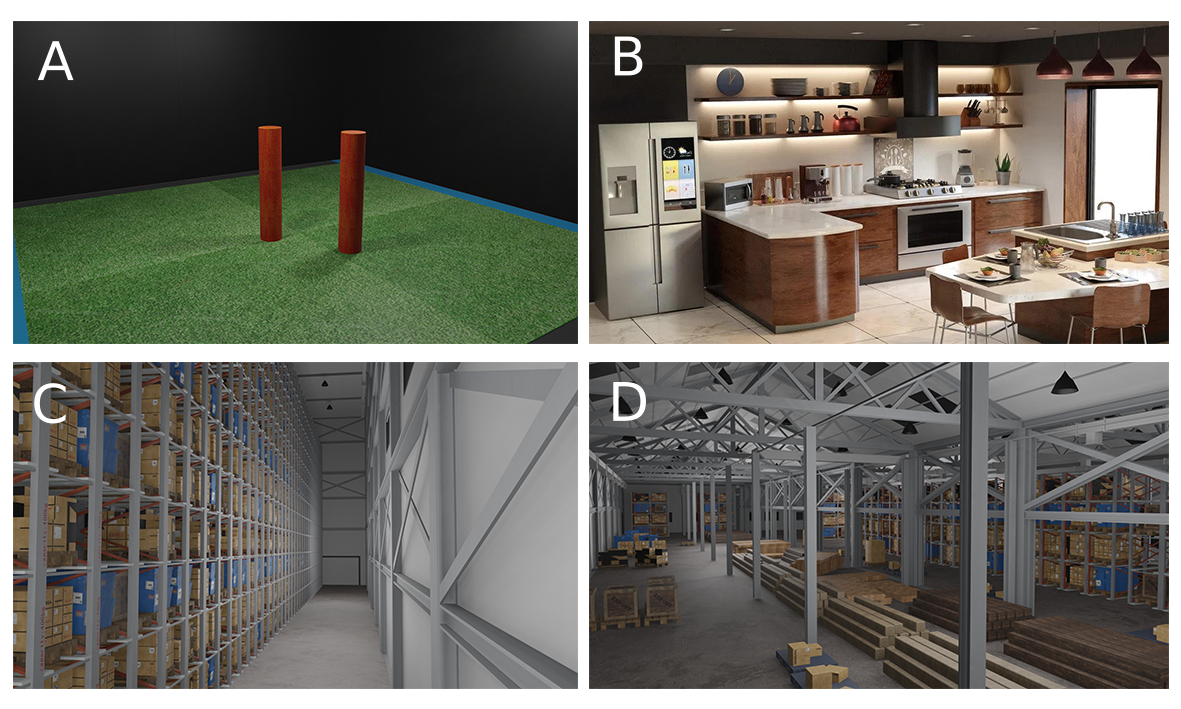}
    \setlength{\belowcaptionskip}{-10pt}
    \caption{Rendering of four scenes used in the simulated benchmark: \textbf{(A)} the TU Delft flying arena, \textbf{(B)} a kitchen, \textbf{(C)} a set of storage shelves, and \textbf{(D)} a wood warehouse.}
    \label{fig:SceneRenders}
\end{figure}

First, the FAITH method is benchmarked in a simulated environment using the ESIM event-based camera simulator~\cite{rebecq2018esim} provided with the DVS240 event-based camera specifications which we used in our indoor obstacle avoidance dataset. Four distinct 3D scenes are exported from the open-source software Blender to \textit{.obj} files. These scenes have different textures and layouts to ensure the diversity of environments (Fig.~\ref{fig:SceneRenders}). We then provide these scenes to the ESIM simulator along with 100 flight trajectories (camera coordinates over time in a \textit{.csv} file). To test the robustness of the methods to different FOE locations, the trajectories are chosen such that the FOE covers all course angles in the FOV ($-30^{\circ}$ to $30^{\circ}$). Both straight trajectories (with different yaw angles) and sway trajectories (varying the FOE during simulation) are used. The ground truth FOE is known from the simulated trajectory and camera pose. 

\begin{figure}
    \centering
    \includegraphics[width=1\linewidth]{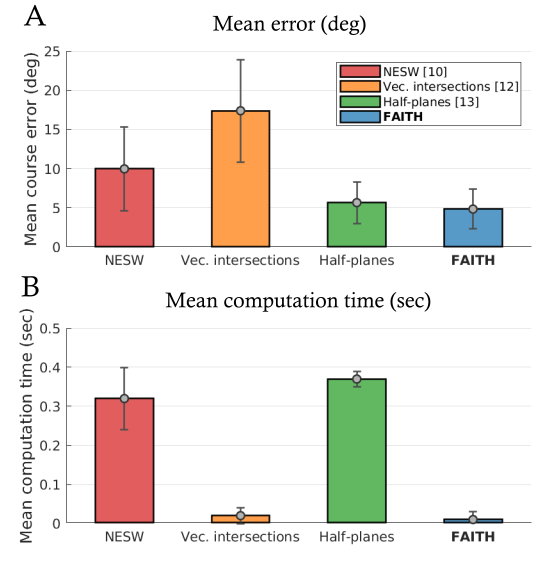}
    \caption{Comparison of the overall performance of our FOE estimation method with three other state-of-the-art methods, after testing over 100 distinct trajectories in the four simulated environments (Fig.~\ref{fig:SceneRenders}). \textbf{(A)} Average angular error (in degrees) in the FOE estimation. \textbf{(B)} Mean computation time (in seconds) required to process $10^3$ OF vectors.}
    \label{fig:MethodsPerformance}
\end{figure}

The results of these $N=100$ simulations are shown in Figure~\ref{fig:MethodsPerformance} and Table~\ref{tab:simRealResults}. The mean course angle estimation error is compared for the four methods. The FAITH method shows state-of-the-art accuracy, with a mean error of $4.84^{\circ}\pm2.53^\circ$. The worst performance is achieved by the 'Vec. Intersections' method with an overall angular error of $17.39^\circ \pm 6.54^\circ$.  Fig.~\ref{fig:MethodsPerformance}-B shows the mean computation time per 1000 vectors. This proves the large reduction in computational effort for our method, confirming the theoretical insight detailed in Section~\ref{sec:compcomplexity}. It also clearly demonstrates the computational efficiency of the 'Vec. intersections' method, which also does not update all probability map pixels and uses a RANSAC scheme. In contrast, the mean course estimation error is significantly larger for the 'Vec. intersections' method, confirming that using the OF vectors instead of half-planes decreases the accuracy (normal vs. real flow).

\begin{table*}[]
    \centering
    \caption{Overall performance obtained with the FAITH method, compared to other state-of-the-art FOE estimation methods, \\for both the simulated benchmark (ESIM) and our obstacle avoidance dataset.}
    \begin{tabular}{ccccc}
        \hline
        \multirow{2}{*}{Method} & \multicolumn{2}{c}{ESIM benchmark ($N=100$)} & \multicolumn{2}{c}{Obstacle avoidance dataset ($N=1300$)} \\ \cline{2-5} 
                           & Angular error           & Computation time & Angular error           & Computation time \\ \hline
        NESW~\cite{Huang2018AnEW}
               & $9.97^\circ \pm 5.34^\circ$ & $0.32 \pm 0.08$ s    & $18.87^\circ \pm 3.95^\circ$ & $0.35 \pm 0.12$ s    \\
        Vec. Intersections~\cite{Buczko2017Monocular}
 & $17.39^\circ \pm 6.54^\circ$ & $0.02 \pm 0.02$ s    & $17.49^\circ \pm 5.41^\circ$ & $0.05 \pm 0.02$ s    \\
        Half-planes~\cite{Clady2014AsynchronousVE}       & $5.66^\circ \pm 2.67^\circ$ & $0.37 \pm 0.02$ s    & $10.60^\circ \pm 3.91^\circ$ & $0.43 \pm 0.15$ s    \\ 
        \textbf{FAITH}         & $\bm{4.84^\circ \pm 2.53^\circ}$ & $\bm{0.01 \pm 0.02}$ s    & $\bm{10.06^\circ \pm 2.88^\circ}$ & $\bm{0.05 \pm 0.02}$ s    \\ \hline
    \end{tabular}
    \label{tab:simRealResults}
\end{table*}

\subsection{Benchmark on an event-based obstacle avoidance dataset}
\label{sec:datasetbenchmark}
\begin{figure}[]
    \centering
    \includegraphics[width=0.9\linewidth]{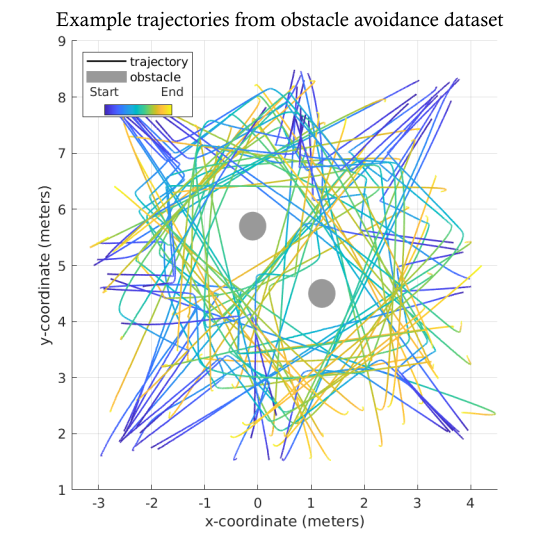}
    \caption{Representation of 78 sample trajectories from the obstacle avoidance dataset. This dataset is used to validate the performance of the FOE estimation method. The MAV is controlled manually and two poles in the center of the TU Delft flying arena are avoided.}
    \label{fig:exampletraj}
\end{figure}

The event-based domain requires new approaches and datasets due to the sparse asynchronous event representation. To address this challenge, a novel obstacle avoidance dataset using a real event-based camera was recorded (see Supplementary Materials). It consists of \hbox{$\sim1350$} manual obstacle avoidance runs performed with an MAV equipped with an event-based camera (DVS240), a 24-GHz radar sensor, a Full-HD RGB camera, a 6-axes IMU, and OptiTrack data for position and attitude ground truth. The obstacles consist of one or two 50-cm wide poles, of which the ground truth location is known (Fig. \ref{fig:exampletraj}). Each trial consists of approximately 10 seconds of recording.

The benchmark on the obstacle avoidance dataset is performed by comparing the four methods. The ground truth FOE is available as the OptiTrack system tracks both the pose and the position of the MAV during the trials. These trials contain a variety of trajectories, obstacles and backgrounds to ensure diversity of environment and motion.

\begin{figure}
    \centering
    \includegraphics[width=0.9\linewidth]{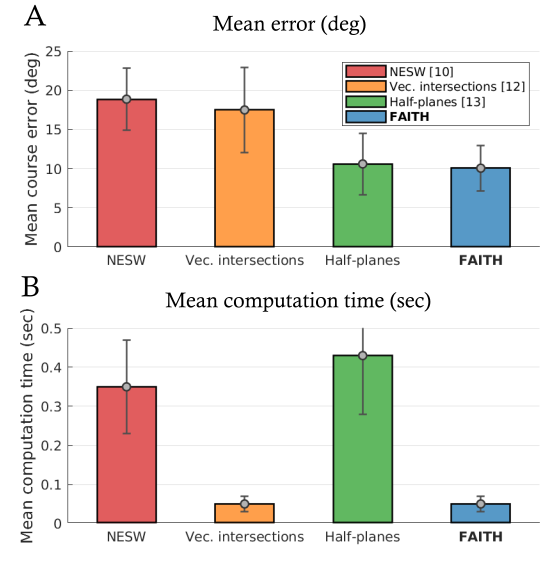}
    \caption{Comparison of the overall performance of the FAITH method with three other state-of-the-art FOE estimation methods, after testing over 1300 samples of our obstacle avoidance dataset. \textbf{(A)} Average angular error (in degrees) in the FOE estimation. \textbf{(B)} Mean computation time (in seconds) required to process $10^3$ OF vectors.}
    \label{fig:realbenchmark}
    \hspace*{-0.5cm}
\end{figure}

The results of this benchmark, as seen in Figure~\ref{fig:realbenchmark} and Table~\ref{tab:simRealResults}, confirm those obtained with the simulator indicating that the FAITH method outperforms the others. Comparing these results show that the FOE estimation accuracy of all methods on the live recorded data is lower than when using simulated data. This is a consequence of multiple factors, such as the higher amount of noise from the DVS240 camera, vibrations caused by the propellers of the MAV or the increased sparsity of the OF due to the texture and lightning conditions of the scene. In contrast, the relative performance of the methods does not change, our method is still the most accurate and computationally efficient of these methods.

\subsection{Experiment onboard MAV}
To show onboard performance of the FAITH method, it is implemented within the ROS (Robot Operating System) framework using C\texttt{++} and used in an autonomous obstacle avoidance task. The MAV is set to fly straight-forward at a constant velocity in the flying arena and encounters a pole approximately halfway. The obstacle avoidance algorithm then detects the pole and gives an avoidance command to the iNav Autopilot running onboard the MAV. 

\subsubsection{Obstacle avoidance strategy}
A straight-forward obstacle avoidance algorithm is designed using OF as input and an avoidance course as output which is fed to the iNav autopilot. In order to detect an obstacle, OF is clustered based on the concatenated image coordinate (normalized between 0~and~1) and TTC (normalized by mean and variance). The FOE is estimated using the FAITH method. The TTC is calculated using this FOE estimation (Appendix~\ref{app:ttc}). To cluster the vectors, we apply a Density-Based Spatial Clustering of Applications with Noise (DBSCAN)~\cite{Ester96adensity-based} with $\epsilon = 0.2$, $\textit{minPts} = 20$ and an Euclidean distance measure. High TTC values are clipped to a user-defined maximum. The mean TTC of the clusters is calculated and the cluster with lowest TTC is assumed to be the highest priority obstacle. A bounding box is drawn around this obstacle cluster. If the FOE location is within the obstacle region and the mean obstacle TTC is below a user-defined threshold, the algorithm gives an 1.5 second roll command to the autopilot to avoid the obstacle. The sign of the roll command is determined by selecting the direction towards the cluster with the highest mean TTC.

\begin{figure}
    \centering
    \includegraphics[width=0.9\linewidth]{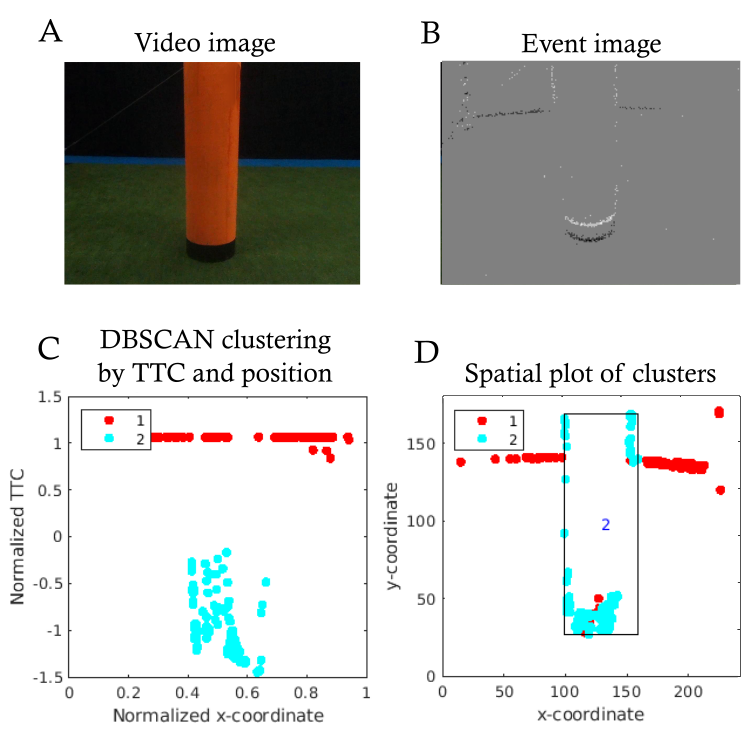}
    \caption{Example of DBSCAN clustering for the onboard FOE estimation experiment, implementing the FAITH method in an obstacle avoidance task.  \textbf{(A)} Frame-based image of the pole. \textbf{(B)}  Event-based image of the pole. \textbf{(C)} Clustering optic flow based on TTC and position.  \textbf{(D)} Clusters, mapped to a spatial plot. As Cluster 2 has the lowest mean TTC, it is identified as (highest priority) object and a bounding box is drawn.}
    \label{fig:ClusteringExample}
\end{figure}

\subsubsection{Hardware architecture} 
\begin{figure}
    \centering
    \includegraphics[width=\linewidth]{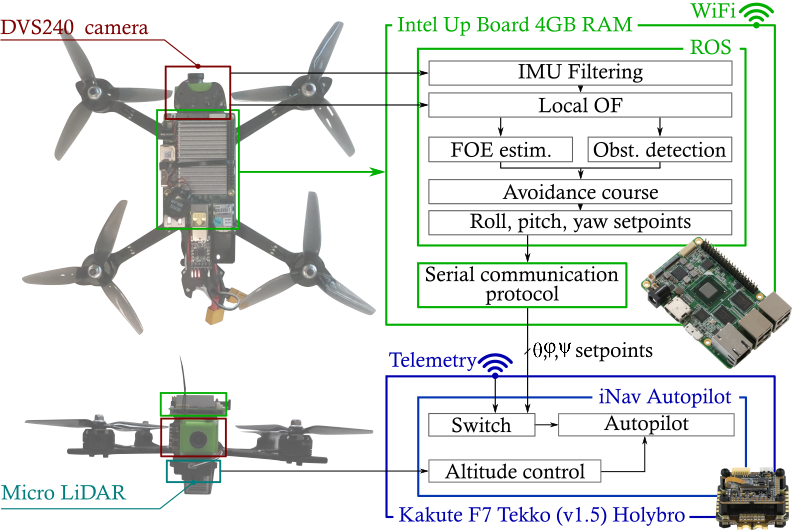}
    \caption{Hardware architecture of the MAV. The visual processing and the obstacle avoidance algorithms are processed onboard Intel's Up board within the ROS environment. A switch allows the user to switch from manual control to autonomous mode. In both cases, the altitude is kept constant by means of the micro Lidar.}
    \label{fig:hardware}
\end{figure}

The MAV is a quadrotor built upon the GEPRC FPV frame kit Mark4, featuring the Kakute F7 Tekko ESC Combo v1.5 flashed with the iNav autopilot. The embedded CPU consists in the Intel Up board (64-bits Intel Atom x5 Z8350 1.92GHz Processor) running the Linux 18.04 LTS operating system. An overview of the hardware architecture is provided in Fig.~\ref{fig:hardware}. The board is used for data acquisition and processing, autonomous navigation, and wireless communication with the host machine. All MAV-related embedded processing, i.e., DVS data acquisition, OF and FOE estimation, obstacle avoidance and navigation, are performed within the ROS (Robot Operating System) framework. As for the obstacle avoidance dataset, the MAV is equipped with the DVS240 event-based camera ($240 \times 180$ pixels). The altitude of the drone is controlled separately with a downward-facing micro LiDAR (TFMini, QWiic). 

\subsubsection{Experimental setup} During the experiments, the MAV ground truth position and attitude are determined by the OptiTrack motion capture system installed in the flying arena. A pole, of which the ground truth location is known, is positioned in the center of the flying arena. The MAV is set to autonomously fly along a straight trajectory, from 12 different starting positions and headings. A set of 60\% of the trajectories are designed as collision courses with the pole, while the remaining 40\% trajectories concern a near pass of the pole. This configuration is meant to qualitatively assess the robustness of the FOE estimation and obstacle avoidance methods in real-world conditions.  

\subsubsection{Results}
\begin{figure}
    \centering
    \includegraphics[width=0.7\linewidth]{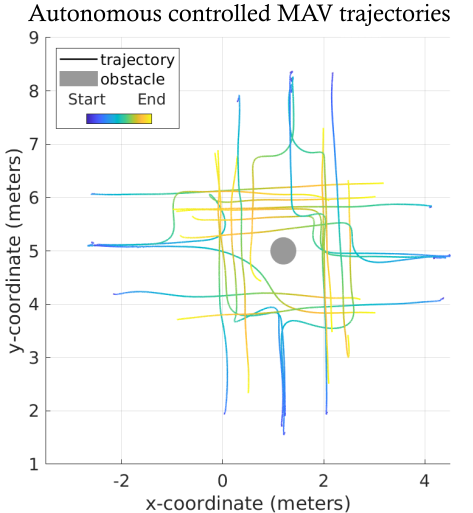}
    \caption{20 Successful autonomous obstacle avoidance trajectories using the FAITH method to estimate the FOE. This shows the successful implementation of our method in an obstacle avoidance task.}
    \label{fig:manualVSauto}
\end{figure}

The autonomous obstacle avoidance method, using FAITH to estimate the FOE, is shown to perform a successful obstacle avoidance manoeuvre in $80\%$ of the runs (20 out of 25). The faulty runs are a result of the low-textured scene, which impedes the FOE estimation. When the potential FOE area (Fig.~\ref{fig:halfplane}) is not fully bounded by OF, the FOE estimation becomes less accurate. This also influences the TTC estimation and subsequently deteriorates the clustering quality. As a result, occasionally when no fully bounding OF is generated in the scene, the object is not detected correctly. Fig.~\ref{fig:manualVSauto} shows the trajectories of 20 successful obstacle avoidance runs. This figure shows the ability of the MAV to autonomously determine its course using the FAITH method and avoid the object. This shows the successful onboard performance of our method in a real-time obstacle avoidance task.

\section{Conclusion and future work}
We introduced the novel FAITH method to determine the course of an MAV by means of an event-based camera, along with a fast RANSAC-based algorithm for the determination of the FOE. Using event-based normal OF as input, the method is able to efficiently estimate the course of the MAV. The accuracy and computational performance are validated by performing a benchmark using both a simulated event-based camera data and a novel live obstacle avoidance dataset containing real sensor data. On both simulated and real event-based camera data, the FAITH method shows a state-of-the-art accuracy, with a beyond state-of-the-art computational performance. 

We further tested our method in an obstacle avoidance task onboard an MAV, successfully demonstrating real-time performance of our method. The limitations of OF-based strategies in low-textured environments show the bottleneck towards MAV autonomous applications, also suggested by results obtained with our dataset. 

\section*{Acknowledgment}
This work is part of the Comp4Drones project and has received funding from the ECSEL Joint Undertaking (JU) under grant agreement No. 826610. The JU receives support from the European Union's Horizon 2020 research and innovation program and Spain, Austria, Belgium, Czech Republic, France, Italy, Latvia, Netherlands.

\section*{Supplementary materials}
\label{sec:supplementary}
The ROS implementation of FAITH can be found here: \url{https://github.com/tudelft/faith}, and the supporting video \url{https://youtu.be/X09mIqoqAFU}. The Obstacle Detection and Avoidance dataset is available at: \url{https://github.com/tudelft/ODA_Dataset}.

\bibliographystyle{IEEEtran}
\bibliography{refs}

\appendices

\section{Optic Flow, Focus of Expansion and Time-To-Contact Theory}
\label{app:OFtheory}
\subsection{Optic Flow Theory}
\begin{figure}
    \centering
    \includegraphics[width=0.4\textwidth]{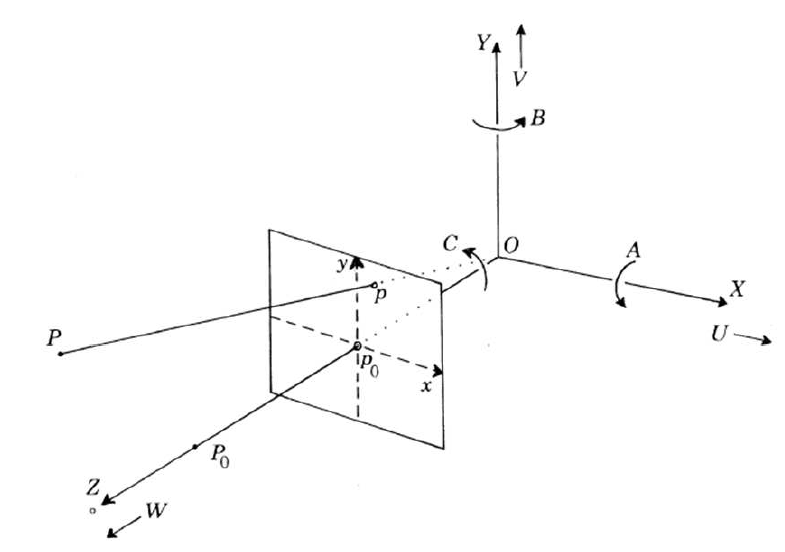}
    \caption{Optic flow reference system. From Longuet-Higgins et al.~\cite{LonguetHiggins1980TheIO}.}
    \label{fig:OFreference}
\end{figure}

Optic flow (OF) consists of two components, due to translation and rotation. The OF generated by translation gives information about the scene and the ego-motion of the observer. In contrast, the OF generated by rotation does not provide any insights on translational ego-motion. Therefore, the OF in this research is derotated using an onboard IMU such that only OF based on translation is used. This research uses an event-surface method, firstly proposed by Benosman et al.~\cite{Benosman2014event}, and later improved for online application by Hordijk et al.~\cite{Hordijk2018vertical}. This method generates sparse normal OF. In order to describe the underlying geometry of OF, an arbitrary point from the 3D world is projected on a 2D surface. The projected point on the surface has the following coordinates (Fig.~\ref{fig:OFreference}).

\begin{equation}
 x = \frac{X}{Z},\quad y = \frac{Y}{Z}
\end{equation}

To determine the motion of this point, the equation above is differentiated with respect to time.

\begin{equation}
\Dot{x} = \frac{\Dot{X}}{Z} - \frac{X\Dot{Z}}{Z^2}, \quad \Dot{y} = \frac{\Dot{Y}}{Z} - \frac{Y\Dot{Z}}{Z^2} 
\end{equation}

Values for $\Dot{X}$, $\Dot{Y}$ and $\Dot{Z}$ can be derived (for a derivation see Longuet-Higgins et al. \cite{LonguetHiggins1980TheIO}), resulting in the following OF equations.

\begin{equation}
\label{eq:OF}
\begin{split}
u = -\frac{U}{Z} + x\frac{W}{Z}+ Axy - Bx^2 - B + Cy = u_T + u_R\\
v = -\frac{V}{Z} + y\frac{W}{Z}- Cx + A + Ay^2 - Bxy = v_T + v_R 
\end{split}
\end{equation}

Note that these equations consist of a translational ($u_T, v_T$) and rotational ($u_R, v_R$) component. The rotational component is a result of camera rotations and does not contain information about the ego-motion of the observer. This effect is compensated in this research by using the known ego-rotation from an onboard Inertial Measurement Unit.

\subsection{Focus of Expansion Theory}
When an observer translates through a static scene, the OF diverges from a singular point called the FOE. At this location on the image, the OF is zero and all OF is directed outwards. This position is an indication of the course of the observer. If the OF is zero and the rotational component is filtered out, the following derivation is made using Eq.~\ref{eq:OF}.

\begin{equation}
\begin{split}
u_T = 0 = -\frac{U}{Z} + x_{FOE}\frac{W}{Z}\\
v_T = 0 = -\frac{V}{Z} + y_{FOE}\frac{W}{Z} 
\end{split}
\label{eq:FOE_1}
\end{equation}

Rewriting these equations gives the following result.

\begin{equation}
x_{FOE} = \frac{U}{W},\quad y_{FOE} = \frac{V}{W}
\label{eq:FOE_2}
\end{equation}

To show OF diverges from the FOE, (\ref{eq:FOE_1}) and (\ref{eq:FOE_2}) are used to re-express $u_T$ and $v_T$.

\begin{equation}
\begin{split}
u_T = -\frac{U}{Z} + \frac{x W}{Z} =  (-\frac{U}{W} + x)\frac{W}{Z} = (x - x_{FOE})\frac{W}{Z}\\
v_T = -\frac{V}{Z} + \frac{y W}{Z} =  (-\frac{V}{W} + y)\frac{W}{Z} = (y - y_{FOE})\frac{W}{Z} 
\end{split}
\label{eq:FOE_3}
\end{equation}

Rewriting this equation shows the geometrical relation which results in the OF diverging from the FOE.

\begin{equation}
\frac{u_T}{v_T} = \frac{x-x_{FOE}}{y-y_{FOE}} 
\label{eq:FOE_4}
\end{equation}

This geometrical relation is used as basis for the methods discussed in the benchmark (Section~\ref{sec:benchmark}).

\subsection{Time-To-Contact Theory}
\label{app:ttc}
The Time-To-Contact (TTC) is a property of each point in an image, describing its relative velocity to the camera principle axis. Eq.~\ref{eq:FOE_3} can be rewritten to the following equation for divergence.

\begin{equation}
\frac{W}{Z} = \frac{u_T}{x-x_{\text{FoE}}} = \frac{v_T}{y-y_{\text{FoE}}} 
\end{equation}

\noindent The divergence is inversely related to the TTC.
\begin{equation}
\tau = \frac{Z}{W}
\end{equation}

In the onboard test of the FAITH method for estimating the FOE, an obstacle detection method is used which clusters the OF based on the vector position and TTC. Divergence is inversely related to the TTC and has converging properties. Although this seems an advantage over TTC, divergence values are much lower (i.e. zero for infinite obstacle distance or zero observer velocity) and unsuitable for proper clustering. Therefore, TTC is chosen as primary clustering variable in this research.

\section{FOE outside the field of view}
\label{app:FOEoutside}

\begin{figure}[H]
    \centering
    \includegraphics[width=0.85\linewidth]{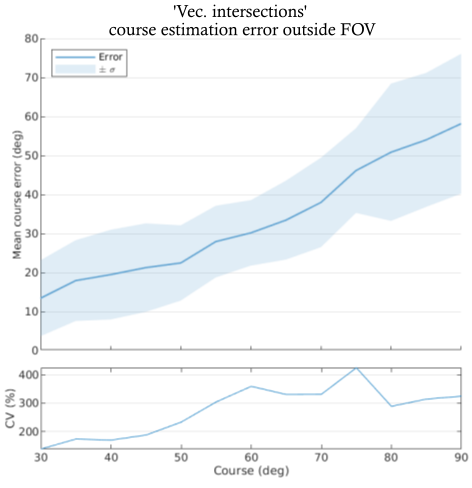}
    \caption{Performance of the 'Vec. intersections' method implemented by Buczko et al.~\cite{Buczko2017Monocular} for a course of $30^{\circ}$ to $90^{\circ}$, outside the FOV. The lower plot shows the coefficient of variation as percentage, $\text{CV}=\frac{\mu}{\sigma}$}
    \label{fig:perfOutsideFOV}
\end{figure}

The performed benchmark on simulated and real event-based camera data considers only FOE locations inside the camera field of view. In the obstacle avoidance strategy, the MAV flies towards clusters which are within the field of view, as this gives certainty about the scene the MAV is flying towards. If the MAV flies a course which is outside the field of view, our method will provide an unbounded FOE region, and thus also no exact FOE location. Although this is a limitation of our method, it does provide the general direction the MAV is moving towards. The side in which the FOE region is unbounded is also side on which the FOE lies, thus the FOE location is bounded to a half-plane. Of the compared methods in the benchmark, only the vector intersections method (e.g. implemented by Buczko et al.~\cite{Buczko2017Monocular}) is able to estimate FOE locations outside the field of view. Fig.~\ref{fig:perfOutsideFOV} shows the performance of the vector intersections method for 40 simulated trials, with an FOE angle ranging from $30^{\circ}$ to $90^{\circ}$. The course estimation error and CV grows rapidly as the course is further outside the FOE. This results in CV values of over $300\%$, which show a very low estimation certainty. Therefore, it is concluded that this method has a limited advantage over our method regarding estimating the FOE outside the field of view.

\end{document}